\pdfoutput=1
\documentclass[letterpaper]{article} % DO NOT CHANGE THIS
\usepackage{aaai23}  % DO NOT CHANGE THIS
\usepackage{times}  % DO NOT CHANGE THIS
\usepackage{helvet}  % DO NOT CHANGE THIS
\usepackage{courier}  % DO NOT CHANGE THIS
\usepackage[hyphens]{url}  % DO NOT CHANGE THIS
\usepackage{graphicx} % DO NOT CHANGE THIS
\usepackage{natbib}  % DO NOT CHANGE THIS AND DO NOT ADD ANY OPTIONS TO IT
\usepackage{caption} % DO NOT CHANGE THIS AND DO NOT ADD ANY OPTIONS TO IT
\usepackage{algorithm}
\usepackage{algorithmic}

\usepackage{newfloat}
\usepackage{listings}
\DeclareCaptionStyle{ruled}{labelfont=normalfont,labelsep=colon,strut=off} % DO NOT CHANGE THIS
\floatstyle{ruled}
\newfloat{listing}{tb}{lst}{}
\floatname{listing}{Listing}
\title{Is it Required? Ranking the Skills Required for a Job-Title}

\author {
    Sarthak Anand,\textsuperscript{\rm 1 2} 
    Jens-Joris Decorte, \textsuperscript{\rm 2 3}
    Niels Lowie \textsuperscript{\rm 2}
}
\affiliations{
    \textsuperscript{\rm 1} Utrecht University\\
    \textsuperscript{\rm 2} TechWolf\\
    \textsuperscript{\rm 3} Ghent University \\
}

\usepackage{bibentry}
\begin{document}

\maketitle

\begin{abstract}
In this paper, we describe our method for ranking the skills required for a given job title. Our analysis shows that important/relevant skills appear more frequently in similar job titles. We train a Language-agnostic BERT Sentence Encoder (LaBSE) model to predict the importance of the skills using weak supervision. We show the model can learn the importance of skills and perform well in other languages. Furthermore, we show how the Inverse Document Frequency factor of skill boosts the specialised skills.
\end{abstract}

\section{Introduction \& Related Work}
Hiring the right candidate can be a challenging and time consuming task. Companies are constantly trying to  optimize this process by building systems to engage with more relevant candidates.  
Knowledge about skills and job titles associated with a job vacancy helps in automating the matching process to a suitable candidate. Recent work in this direction to explore the capabilities of BERT to learn the representations of the job titles was done by \cite{job-bert}. Following this work,
\cite{bhola-etal-2020-retrieving} worked on identifying the skills from the job descriptions through an eXtreme Multi-Classification (XLMC) objective. One drawback of treating the task as a classification problem is that some skills are more important or relevant for a job vacancy than others. For instance for a job of \textbf{Python Web Developer} knowledge of \textit{Python} would be more relevant than \textit{Problem Solving} or \textit{Communication Skills}. To solve this,
we frame the problem to learn the importances of skills given a job title. We only take job input into account in the , since the complete description can be noisy and contain irrelevant information as well.

\section{Dataset \& Methodology }
\subsubsection{Dataset} We use an in-house taxonomy of around 37K skills, which was collected using a mix of sources and data mining techniques. We collect job postings from various job portals, and using the skills taxonomy we collect job titles and the skills associated with them. We also do some pre-processing to clean the job titles. In the end, we have around 170K unique job titles in English with a list of skills associated with them, which we later split into training, development and test in 70:10:20 ratio respectively.

\subsubsection{Weak Supervision} 
\begin{figure}
    \includegraphics[width=1.0\linewidth]{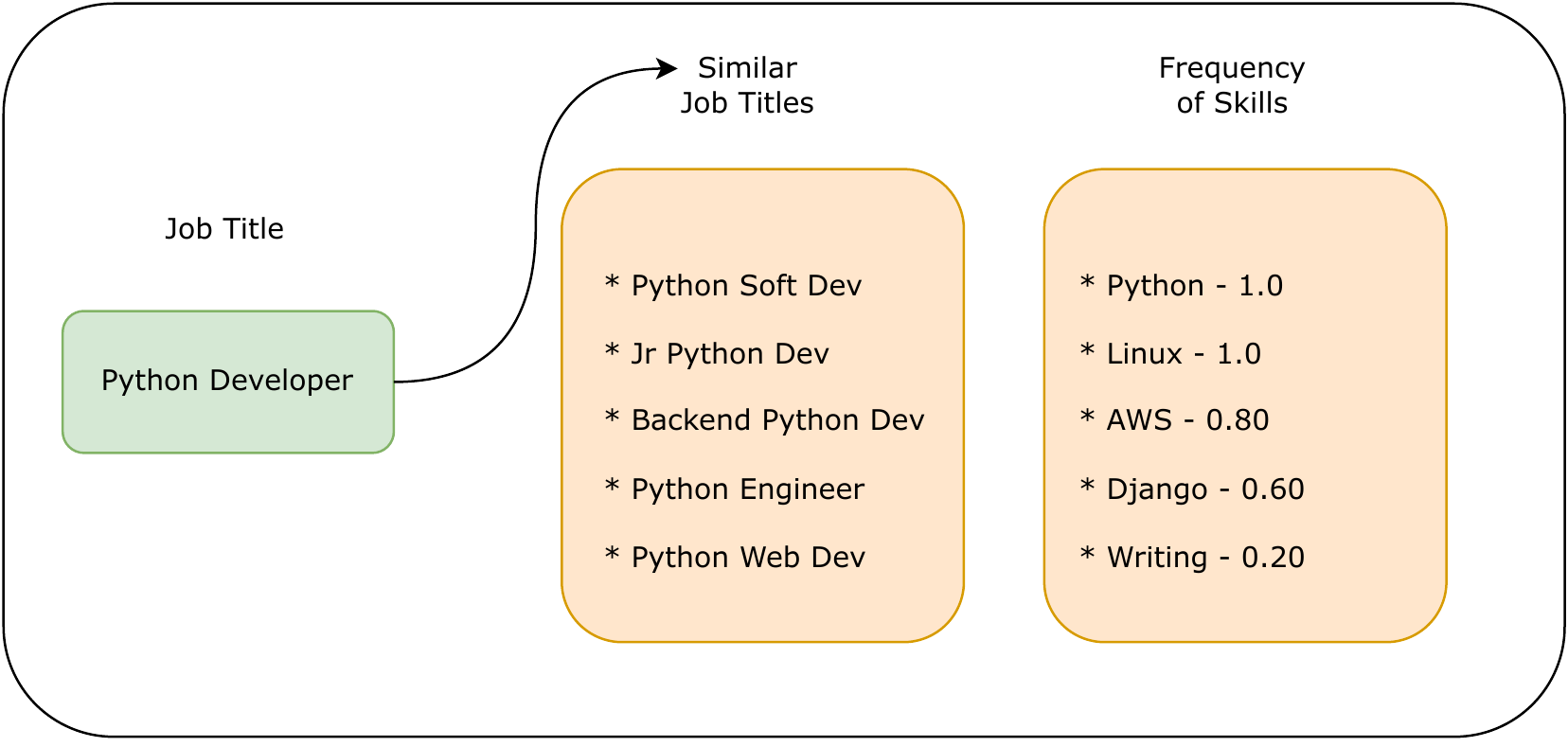}
    \caption{In the case of \textit{Python Developer} we find similar job titles in the training dataset (center). Secondly, we then find the relative frequency of skills within the similar job titles (right). The model is then trained to predict these frequencies directly from the job title.}
    \label{data}
\end{figure}

Similar job positions can be represented in multiple ways, for instance \textit{Python Developer} can also be expressed as \textit{Python Programmer, Python Software Engineer} etc. In our analysis, the skills that appear more frequently in similar job titles tend to be more important and relevant for a given title. Therefore, for each job title we first find similar job titles using the sentence embedding of the job titles using Sentence BERT \cite{reimers-2019-sentence-bert}. The threshold for the cosine similarity is set to 0.75 after empirical analysis. We then find the frequency of all skills present within the similar job titles and normalise it on a scale of 0-1, which are then learned by the model to predict for the input job title. This process is illustrated in the Figure \ref{data}.

\subsubsection{Training}
\begin{figure}
    \includegraphics[width=1.0\linewidth]{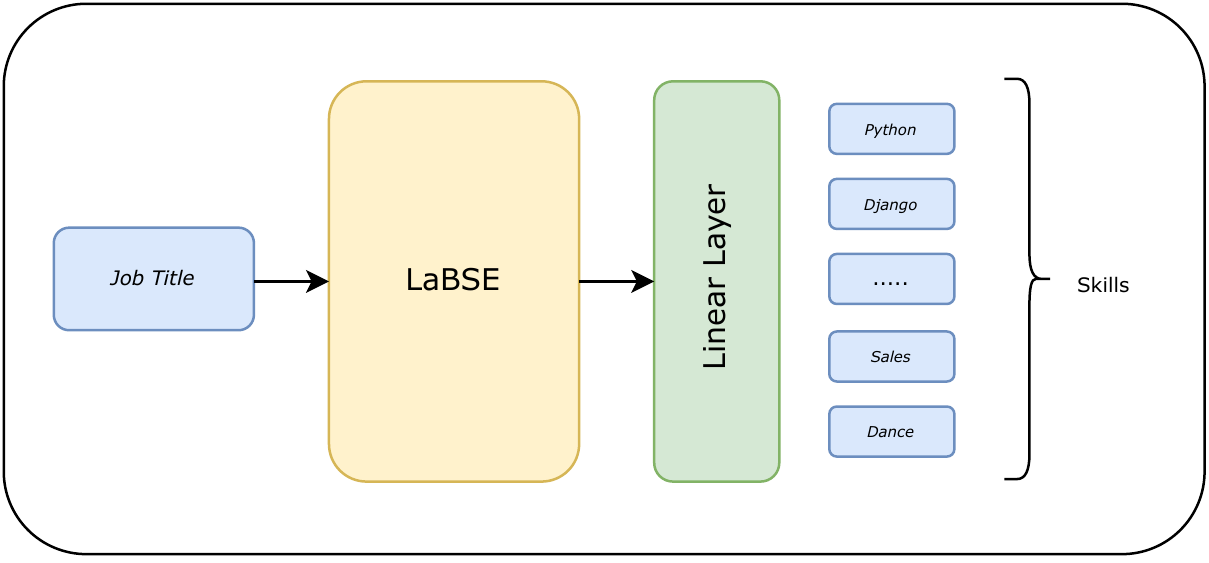}
    \caption{Our Model predicts the importance of each skill which is represented by nodes in the linear layer.}
    \label{model}
\end{figure}

The training objective of the model is to predict the importance of a skill given a job title. We use the pre-trained Language-agnostic BERT sentence embedding (LaBSE) model \cite{feng-etal-2022-sbert} to extract the sentence encoding of the job title. The encoding is then passed to a linear layer, and sigmoid activation is applied to map the output values between 0 and 1, as illustrated in Figure \ref{model}.
Finally, we rank the skills for the job titles using the computed importance.

\section{Results \& Discussion}
\subsubsection{Results} In order to evaluate the models during training we use Mean Average Precision for top 20 skills. The table \ref{results-metrics} shows the improvement on fine tuning the pre-trained LaBSE model. The Table \ref{result-1} shows the results for different job titles. The model also performs fairly on other languages (as seen in the table \ref{result-multi}), but by fine tuning we lose the multi-lingual capabilities of the ranking system.

\subsubsection{Boosting specialised skills} If we look at the top skills for \textit{Stock Broker} in the Table \ref{result-1}, we find some skills which are very generic and therefore have high occurrence in most of the job titles like \textit{Excel, Communication Skills, Sales, Finance}. In-order to find specialised skills we apply a technique similar to TF-IDF, using the training data we compute the $IDF_s$ score of each skill $s$ as \\
%\[ IDF(S) =  log(\frac{N}{f_s }) \]
\[ IDF_s =  log(\frac{N}{f_s}) \]
where $N$ equals the total number of titles and $f_s$ is the number of titles associated with skill $s$.
On comparing the IDF scores for \textit{Excel (0.92)} and \textit{Tableau (4.72)}, we get an idea how it helps specialised skills. For finally ranking the skills we compute the importance of skill from the model and multiply with the IDF score of the skill.

The improvement in the rankings can be easily seen in table \ref{result-idf}.

\begin{table}[]
    \centering
    \begin{tabular}{c|c}
        \textbf{Model} &  \textbf{Mean AP@20}\\
        \hline
        Pre-trained LaBSE & 0.621 \\
        Fine-tuning LaBSE & 0.722 \\
        \hline
    \end{tabular}
    \caption{Evaluation of the Models on the Test Set}
    \label{results-metrics}
\end{table}

\begin{table}[]
    \scalebox{0.8}{
    \centering
    \begin{tabular}{c|c|c}

       \textbf{Front-End Dev} & \textbf{Stock Broker} & \textbf{Therapist} \\
        \hline
        JavaScript & Sales & Therapy \\
        Software Development & Communication Skills & Healthcare \\
        Java & Marketing & Patient Care\\
        CSS & Customer Service & Nursing \\
        AngularJS & Excel & Group Therapy\\
        Web Applications & Retail & Mental Health \\
        JQuery UI & Finance & Sociology \\
        \hline
    \end{tabular}}
    \caption{Ranking of the skills for given job-title}
    \label{result-1}
\end{table}

\begin{table}[]
    \centering
    \scalebox{0.8}{
    \begin{tabular}{c|c|c}
        \textbf{Doctor} & \textbf{Arzt} & \textbf{Docteur} \\
        \hline
        CME & Family Medicine & Family Medicine \\
        Family Medicince & CME & CME \\
        Emergency Medicine & Emergency Medicine & Emergency Medicine\\
        EMR & EMR & EMR \\
        Urgent Care & Patient Care & Patient Care \\
        \hline
    \end{tabular}}
    \caption{Top Skills for job titles in different languages: Doctor (Eng), Arzt (Ger), Docteur (Fre) for the pre-trained LaBSE model}
    \label{result-multi}
\end{table}

\begin{table}[]
    \scalebox{0.9}{
    \centering
    \begin{tabular}{c|c}

       \textbf{Stock Broker ( without IDF )} & \textbf{Stock Broker ( with IDF )} \\
        \hline
        Sales & Financial Markets\\
        Communication Skills & Securities \\
        Marketing & Equities \\
        Customer Service & Proprietary Trading \\
        Excel & Commodities\\
        Retail & Financial Services \\
        Finance & Hedging \\
        \hline
    \end{tabular}}
    \caption{Ranking of skills without (left) and with (right) IDF weighing}
    \label{result-idf}
\end{table}

\section{Conclusion \& Future Work}
We address the problem of ranking of skills for a given job-title. Importantly, we propose a novel method that does not require manually annotating the importance of skills for different job titles. Further, we show how IDF factor of a skill can improve the rank of specialised skills. In future work can be done to rank skills required using the entire job description.

%\nobibliography{yourbibfile} 
\bibliography{ref}

\end{document}